\documentclass{article}

\usepackage{PRIMEarxiv}

\usepackage[utf8]{inputenc} 
\usepackage[T1]{fontenc}    
\usepackage{hyperref}       
\usepackage{url}            
\usepackage{booktabs}       
\usepackage{amsfonts}       
\usepackage{nicefrac}       
\usepackage{microtype}      
\usepackage{lipsum}
\usepackage{fancyhdr}       
\usepackage{graphicx}       
\graphicspath{{media/}}     

\usepackage{mathtools} 
\usepackage{booktabs} 
\usepackage{tikz} 

\usepackage{pgfplots}
\pgfplotsset{compat=1.18} 
\usepgfplotslibrary{groupplots}

\usepackage{gensymb}
\usepackage{amssymb}
\usepackage{graphicx}
\usepackage{subfig} 
\usepackage{adjustbox} 
\usepackage{natbib} 
\title{Efficient Post-Hoc Uncertainty Calibration via Variance-Based Smoothing}

\author{
    Fabian~Denoodt \\
    University~of~Antwerp \\
    \texttt{fabian.denoodt@uantwerpen.be} \\
    \And
    Jos{\'e}~Oramas\\
    University~of~Antwerp \\
    \texttt{jose.oramas@uantwerpen.be}
}

\begin{document}
    \maketitle
    \begin{abstract}
    Since state-of-the-art uncertainty estimation methods are often computationally demanding, we investigate whether incorporating prior information can improve uncertainty estimates in conventional deep neural networks. Our focus is on machine learning tasks where meaningful predictions can be made from sub-parts of the input. For example, in speaker classification, the speech waveform can be divided into sequential patches, each containing information about the same speaker. We observe that the variance between sub-predictions serves as a reliable proxy for uncertainty in such settings. Our proposed variance-based scaling framework produces competitive uncertainty estimates in classification while being less computationally demanding and allowing for integration as a post-hoc calibration tool. This approach also leads to a simple extension of deep ensembles, improving the expressiveness of their predicted distributions.

    The code to replicate this work is available via GitHub\footnote{\url{https://github.com/anonymoususerforpeerreview/Variance-Based-Softmax-Scaling}}.
\end{abstract}

    \begin{figure}[!htb]
    \centering
    \includegraphics[width=0.8\textwidth, trim=45 5 40 30, clip]{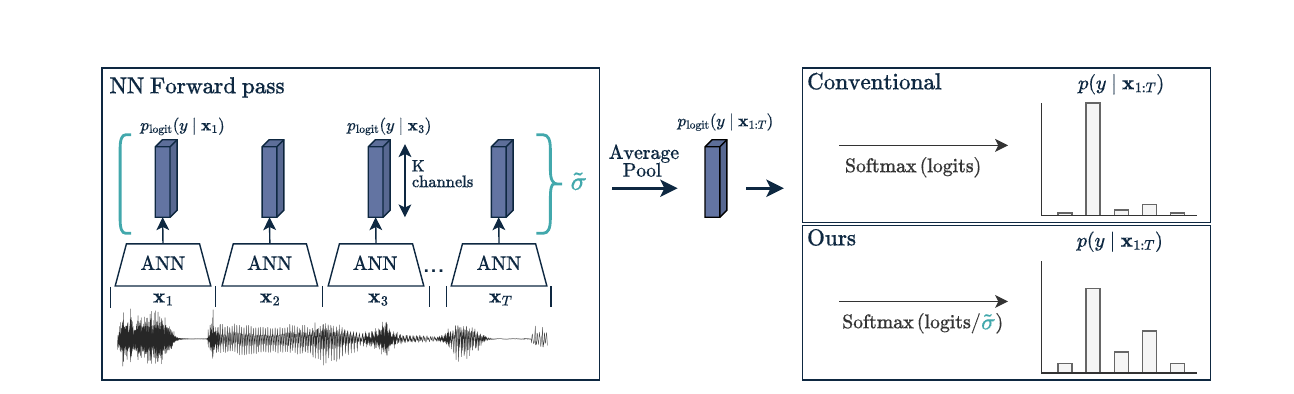}
    \caption{Overview of our method. We use a pre-trained classifier and variance between logits of sub-patches as an uncertainty proxy to recalibrate uncertainty estimates.}
    \label{fig:overview}
\end{figure}

\section{Introduction}\label{sec:intro}
Deep neural networks (NNs) have achieved remarkable performance across a wide range of machine learning tasks and are increasingly used in critical applications~\citep{krizhevsky2012imagenet, ioffe2015batch, miotto2018deep}. However, while they excel in predictive accuracy, they often struggle to quantify uncertainty reliably, leading to overconfident predictions~\citep{DBLP:conf/nips/Lakshminarayanan17, liu_uncertainty_via_distance_awareness, pmlr-v70-guo17a}. This is a serious problem in high-stakes domains such as healthcare and autonomous systems, where well-calibrated confidence estimates are crucial. A perfectly calibrated model, for example, would predict ``cat'' with 70\% confidence and be correct 70\% of the time.

State-of-the-art methods for improving uncertainty estimation, such as deep ensembles, Bayesian neural networks, and MC-dropout~\citep{DBLP:conf/nips/Lakshminarayanan17, blundell2015weight, DBLP:conf/icml/GalG16}, have shown strong results. However, these methods come with substantial computational costs. Ensembles require training multiple models, increasing both memory usage and inference time. Bayesian neural networks, which learn distributions over weights, suffer from slow training and expensive inference. MC-dropout, while lighter than ensembles, still requires multiple stochastic forward passes during inference and keeping dropout active at test time, making it less suitable for real-time applications. These constraints limit the practicality of existing uncertainty quantification (UQ) methods in resource-constrained environments, such as autonomous drones or embedded systems.

Interestingly, many real-world tasks involve input data with inherent redundancy, where the problem can be solved using substructures or subsets of the input.
For instance, in speech classification, short audio patches often contain enough information to determine the speaker. Similarly, in radio signal identification, brief signal windows can be sufficient to identify the transmitting device. Despite this, existing UQ methods do not explicitly exploit such redundancy, leaving an opportunity for more efficient approaches.

In this work, we propose a novel variance-based softmax scaling approach to recalibrate a model's uncertainty estimates by exploiting tasks that follow what we refer to as the ``informative sub-patches'' assumption. Specifically, we introduce a method that uses the variance between predictions on sub-patches of the input as a proxy for uncertainty, achieving real-time inference speeds while improving the model's reliability. We refer to this approach as Variance-based Smoothing (Fig.~\ref{fig:overview}).

Our contributions are threefold:
\begin{enumerate}
    \item We propose a simple yet effective recalibration method that serves as a post-hoc uncertainty estimation technique for pre-trained models while preserving their original accuracy.
    \item We demonstrate competitive uncertainty estimates on both clean and noisy data for diverse tasks, including speech, vision, and radio signal classification while being significantly more efficient.
    \item Our proposal also leads to a simple extension of ensembles, improving upon the expressiveness of their predicted distributions in tasks with many classes while eliminating the need for the informative sub-patches assumption.
\end{enumerate}

    \section{Exploiting Substructure for Uncertainty Estimation}
This section is structured as follows: we begin with a naive approach to exploiting this \textit{informative sub-patches assumption} and discuss its limitations, followed by a refined method that preserves classification performance. Finally, we consider a generalization of this approach as an extension of ensembles, allowing us to omit this data requirement.

\subsection{Ensemble Models as a Starting Point: A Naive Approach}
Ensemble-based methods train multiple models with different weight initializations. When the same input is passed through these models, their predictions tend to agree on in-distribution data similar to what they were trained on. However, for out-of-distribution data, predictions may vary due to differences in the learned model weights. The final classification output is computed as the mean of the softmax predictions:
\begin{equation}
    p(y\mid \mathbf{x}) = \frac{1}{M}\sum_{m=1}^M p_{\theta_m}(y \mid \mathbf{x}) , \label{eq:ensemb_pred}
\end{equation}
where $p(y\mid \mathbf{x})$ is a probability vector over $K$ classes, $M$ is the number of models, and $\theta_m$ denotes the parameters of the $m$-th model~\citep{DBLP:conf/nips/Lakshminarayanan17}. The predictive distribution also serves as the uncertainty estimate.

Based on these ensembles, a naive single-model alternative that exploits the \textit{informative sub-patches assumption} is to split an input $\mathbf{x}$ into $T$ (non)overlapping sub-patches $\mathbf{x}_1, \mathbf{x}_2 \dots \mathbf{x}_T$ and make separate predictions on each sub-patch individually. The final prediction is obtained by averaging the softmax distributions:
\begin{equation}
    p(y\mid \mathbf{x}) = \frac{1}{T} \sum_{t=1}^{T}   p(y \mid \mathbf{x}_{t}) .
\end{equation}
Unlike conventional NNs, where logits are pooled before applying softmax, this approach averages already-softmaxed outputs. While this can temper overconfident predictions, it also harms predictive accuracy, as each prediction is based on a sub-patch with a limited context window. Additionally, if a given sub-patch $\mathbf{x}_t$ contains little information (e.g., a 500ms audio segment where the speaker is silent), the model may output unconfident logit scores where all class values are negative. After applying softmax, a peak is still ``forced,'' causing the prediction from that sub-patch to contribute equally to those from more informative patches. We found that this method tends to produce overly unconfident predictions for both in-distribution and out-of-distribution data, resulting in less informative output distributions.

\subsection{Variance-based Smoothing} \label{sec:method/var_based_soft_scal}
To address these issues, we preserve the conventional pre-softmax classification output, denoted as $p_\text{logit}(y  \mid \mathbf{x})$, and adjust softmax confidence using temperature scaling. The scaling factor $\tilde\sigma \geq 1$, derived from the variance of sub-predictions, adjusts the softmax function to better account for uncertainty:
\begin{equation}
    p(y\mid \mathbf{x}) = \text{Softmax}\left(\frac{p_\text{logit}(y \mid \mathbf{x})}{\tilde\sigma}\right).
\end{equation}
Since temperature scaling does not affect the predicted class ranking, classification accuracy remains unchanged.

To obtain $\tilde{\sigma}$, the standard deviation $\sigma_k$ is computed across sub-predictions for class $k$ over different sub-patch-indices $t$:
\begin{equation}
    \sigma_k:=\sqrt{\frac{1}{T-1} \sum_{t=1}^T\left(\left[p_{\text {logit }}\left(y \mid \mathbf{x}_t\right)\right]_k-\mu_k\right)^2}, \label{eq:sigmak_var_based_softmax_scaling}
\end{equation}
where $\mu_k=\frac{1}{T} \sum_{t=1}^T\left[p_{\text {logit }}\left(y \mid \mathbf{x}_t\right)\right]_k$. We aggregate the per-class standard deviations ${\sigma_1, \dots, \sigma_K}$ into a single scalar $\bar \sigma$ and compute $\tilde \sigma$ as follows:
\begin{align}
    \label{eq:sigma_bar_and_sigma_tilde}
    &\bar \sigma = \frac{1}{K} \sum_{k=1}^K \sigma_k, \
    &\tilde \sigma = \max(\alpha (\bar\sigma + \beta), 1).
\end{align}
The $\max$ operation ensures that $\tilde{\sigma}$ never falls below 1, preventing unintended confidence amplification. The hyperparameters $\alpha$ and $\beta$ control the sensitivity of scaling. We found empirically that reasonable values for $\beta$ are between the negative 50th and 95th percentiles of $\bar{\sigma}$ computed over a validation set, ensuring that smoothing activates only beyond a certain threshold. The parameter $\alpha$ determines the strength of the adjustment, with $\alpha=1$ providing balanced uncertainty estimates. Higher values can be used to encourage more conservative or pessimistic predictions, with values between 1 and 5 being reasonable choices.

Overall, this Variance-based Smoothing approach requires only a single model and a single forward pass over the entire input, significantly reducing computational costs compared to ensembles or MC-dropout approaches. The additional variance computation is minimal, making it practical for real-time applications and useful as a post-hoc calibration tool for existing models. The variance computation, however, requires the architecture to preserve spatial information in the final logits, which typically many convolution-based architectures do.

\subsection{Variance-based Smoothing as an Extension of Ensembles} \label{sec:meth_ensemb_extension}
The proposed Variance-based Smoothing procedure in Sec.~\ref{sec:method/var_based_soft_scal} requires a variance source, which it obtains from sub-predictions across \textit{informative} sub-patches. However, this data-requirement and potential restrictions to the architecture can be omitted when an alternative source of variance is available. One such source is the variance between ensemble predictions, specifically, their logits.

In this case, the temperature value $\tilde{\sigma}$ is obtained by averaging and rescaling $\sigma_k$, now defined as the standard deviation of the predicted logits across the models for class $k$:
\begin{equation}
    \sigma_k:=\sqrt{\frac{1}{M-1} \sum_{m=1}^M\left(\left[p_{\text {logit}~\theta_m}\left(y \mid \mathbf{x}\right)\right]_k-\mu_k\right)^2},
\end{equation}
where $\mu_k = \frac{1}{M}\sum_{m=1}^M\left[ p_{\text{logit}~\theta_m}\left( y\mid \mathbf{x}\right)\right]_k$. The final distribution is then computed as:
\begin{equation}
    p(y\mid \mathbf{x}) = \text{Softmax}\left(\frac{ \frac{1}{M}\sum_m^M{ p_{\text{logit}~\theta_m~}(y \mid \mathbf{x})}}{\tilde\sigma}\right).
\end{equation}
This formulation extends ensemble methods while better preserving relative class importance within a single model. Conventional ensembles apply softmax independently to each model’s predictions, often producing distributions with sharp peaks that may obscure relative confidence levels~\citep{DBLP:journals/corr/HintonVD15, DBLP:conf/cvpr/ParkKLC19, DBLP:conf/nips/Lakshminarayanan17}. Additionally, when the number of classes is large, a small ensemble may be unable to represent certain distributions such as the uniform distribution.

In contrast, integrating Variance-based Smoothing extends the expressiveness of ensemble distributions, allowing them to capture a wider range of entropy levels (demonstrated in Sec.~\ref{sec:experim_var_based_smooth_ensemb}) while better preserving the relative class probabilities learned by individual models.

    \section{Experiments}
Our experiments are designed to answer the following research questions: (1) How does Variance-based Smoothing affect calibration and confidence under dataset shifts for datasets that satisfy the informative sub-patches assumption versus a dataset with a weaker form? (2) How do its computational costs compare against related methods? (3) How are uncertainty distributions represented when applied to ensembles with a high number of class labels?

\subsection{Setup: Datasets and Protocol} \label{sec:experim_setup}
We evaluate our method on three publicly available datasets. Two satisfy the \textit{informative sub-patches assumption}, such that their inputs can be split into sub-patches that preserve sufficient task-relevant information. The first is the Radio Signals dataset~\citep{NhemDWPOP25} with 20 class labels, consisting of the ID of the device that emitted the signal. The second is the 100-hour LibriSpeech subset by \citet{DBLP:conf/nips/LoweOV19}, consisting of spoken utterances from 251 speakers, based on~\citet{DBLP:conf/icassp/PanayotovCPK15}. Lastly, we include the CIFAR-10 dataset~\citep{CIFAR10}, which contains a much weaker form of this sub-patch assumption; e.g., based on the upper-left patch in the image, it is much more difficult to make an informed prediction due to insufficient context. Model architectures and training protocols follow those in the respective dataset papers (Appendix~\ref{apdx:experimental_setup}).

\textbf{Baselines}: We compare our lightweight post-hoc method against single-model uncertainty estimation approaches. Specifically, we consider a conventional classifier, MC-dropout~\citep{DBLP:conf/icml/GalG16}, which applies dropout ($p=0.5$) after every ReLU during both training and inference, and Temperature Scaling~\citep{pmlr-v70-guo17a}, a post-hoc method that learns a fixed temperature scalar from the validation set for improved calibration. Both post-hoc methods reuse the weights from the conventional model. Comparisons with ensembles are omitted due to their significantly higher computational cost, making them an unlikely competitor in our setting. Details in Appendix~\ref{apdx:baselines}.

\textbf{Input Splitting and Sub-Patch Processing}: The input $\mathbf{x}$ is implicitly split into sub-patches $\mathbf{x}_1, \dots, \mathbf{x}_T$ based on the ConvNet's architecture. Convolutional layers with specific kernel sizes and strides control the context size and overlap of these sub-patches. We capture logits $\mathbf{z} \in \mathbb{R}^{T \times K}$ before the final pooling layer (architecture details are discussed in Appendix~\ref{apdx:experimental_setup}). Each $\mathbf{z}_t \in \mathbb{R}^{K}$ represents a prediction for sub-patch $\mathbf{x}_t$, previously denoted as $p_{\text{logit}}(y \mid \mathbf{x}_t)$ in Eq. \ref{eq:sigmak_var_based_softmax_scaling}.

To ensure sufficient context, we ``merge'' neighboring logits $\mathbf{z}_t \dots \mathbf{z}_{t+j}$ by applying average pooling across $\mathbf{z}$ with a specific kernel size where necessary. This process ensures that each $\mathbf{z}_t$ captures a broader spatial or temporal region. The variance scalar $\tilde{\sigma}$ is then computed based on these pooled logits $\mathbf{z}_1 \dots \mathbf{z}_{T'}$ rather than the original logits. For Radio Signals, the input $\mathbf{x} \in \mathbb{R}^{1580 \times 2}$ is reduced to $\mathbf{z} \in \mathbb{R}^{79 \times 20}$. A window-pooling operation (kernel size 10, stride 1) is applied, resulting in 70 pooled logits per sample. For LibriSpeech, the input $\mathbf{x} \in \mathbb{R}^{20480 \times 1}$ is reduced to $\mathbf{z} \in \mathbb{R}^{128 \times 251}$, with average pooling (kernel size 40) producing a $\mathbb{R}^{89 \times 251}$ pooled tensor. For CIFAR-10, the input $\mathbf{x} \in \mathbb{R}^{32 \times 32 \times 3}$ is reduced to $\mathbf{z} \in \mathbb{R}^{8 \times 8 \times 10}$, where standard deviations are computed across the 64 logits without additional pooling.

The values of $\alpha$ and $\beta$ are tuned per dataset, balancing entropy levels across different noise levels while maintaining reliable calibration (Sec.~\ref{sec:experim_calibr_and_datasetshift}).
For Radio Signals and CIFAR-10, we set $\alpha=1$ and $\beta=\text{Avg}(\bar{\sigma}) + 0.5$ where $\text{Avg}(\bar{\sigma})$ is the mean of all $\bar{\sigma}$ values computed on the validation set. For LibriSpeech, we use $\alpha=5$ and set $\beta$ to the negative 95th percentile of $\bar{\sigma}$ values.

\subsection{Calibration and Dataset Shift} \label{sec:experim_calibr_and_datasetshift}

\subsubsection{Reliability Diagrams on Clean Samples}
\begin{figure*}[ht]
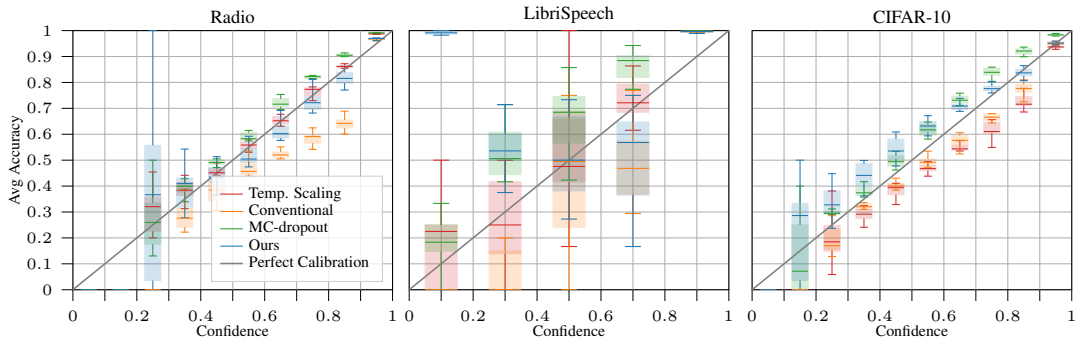

    \centering
    \input{graphs/radio/reliabil_diagr/reliabil_merged}
    \hspace{-1.5em}%
    \input{graphs/libri/reliabil_diagr/reliabil_merged}
    \hspace{-1.5em}
    \input{graphs/cifar/reliabil_diagr/reliabil_merged}
    \caption{Radio, Librispeech, and CIFAR-10: Reliability diagrams on a clean test set. Variance-based Smoothing improves the calibration of the conventional NN on the RADIO and CIFAR-10 datasets. Individual plots are available in Appendix~\ref{sec:apdx_reliabil_diagrs_and_conf_counts}.}
    \label{fig:reliabil_diagr_radio_libri_cifar}
\end{figure*}

To evaluate model calibration, we follow the approach of \citet{pmlr-v70-guo17a}, computing reliability diagrams based on a clean test set. Each prediction is assigned to one of ten confidence bins based on the probability score of their most likely class. The accuracy is then computed for all predictions within a bin. The bins are split into intervals of 10\%: $[0\%, 10\%), [10\%, 20\%) \dots [90\%, 100\%)$, apart from LibriSpeech where bin sizes are intervals of 20\% due to an otherwise insufficient amount of data points per bin. A perfectly calibrated model would produce the identity line.

Fig.~\ref{fig:reliabil_diagr_radio_libri_cifar} displays the results for each dataset. For the Radio and CIFAR-10 datasets, all four models follow the identity line, with the conventional NN consistently being overconfident, particularly in the radio's $[80\%,90\%)$ confidence bin, where the actual accuracy is approximately 65\%. MC-dropout generally produces more calibrated results than conventional but is consistently underconfident (as most of its box plots are above the perfect calibration line). Both Temperature Scaling and Variance-based Smoothing seem to significantly improve calibration on Radio, bringing the conventional NN's predictions closer to the identity line. Interestingly for CIFAR-10, Temperature Scaling seems to sharpen distributions rather than smoothening them, resulting in increased confidence and worse calibration scores. Overall, Variance-based Smoothing offers very competitive calibration scores on both RADIO and CIFAR-10.

For LibriSpeech, calibration follows a similar trend: the conventional model tends to be overconfident, MC-dropout is underconfident, and Variance-based Smoothing falls in between, with box plots both above and below the perfect calibration line. Temperature Scaling performs overall closest to the perfect calibration line. However, all four methods produce wider accuracy variations across runs further away from the perfect calibration line, stressing the difficulty of this 251-class dataset.

Variance-based Smoothing produces some improvement in calibrating the conventional NN, with the 50th percentile of its box plots generally closer to the calibration line, though the effect is less pronounced than in the other datasets. Nonetheless, in the LibriSpeech setting, its value is somewhat limited as the method produces many unconfident but perfectly correct predictions, as indicated by the high empirical accuracy in the $[0, 10\%)$ bin, resulting in overly pessimistic predictions within this specific confidence range.

More detailed figures including the graphs on the number of samples produced in each bin and individual per-method reliability diagrams are available in Appendix~\ref{sec:apdx_reliabil_diagrs_and_conf_counts}.

\subsubsection{Confidence under Dataset Shift}
Following a modified version of \citet{DBLP:conf/nips/SnoekOFLNSDRN19}, we study how confidence evolves under increasing noise perturbations. We consider Gaussian and Speckle noise for Radio and LibriSpeech, while CIFAR-10 is tested under Gaussian, distortion, and affine transformation noise (details in Appendix~\ref{apdx:different_noise_types}). This section first evaluates variance as a proxy for uncertainty in the context of Variance-based Smoothing, followed by an evaluation of confidence across all methods.

\textbf{Variance as a Proxy for Predictive Uncertainty}:
To empirically validate whether the standard deviation across sub-patch predictions ($\bar{\sigma}$) serves as a reliable metric for uncertainty, we incrementally introduce noise and observe how $\bar{\sigma}$ changes. A strong correlation is important, as it is the key informative factor in Variance-based Smoothing.

\begin{figure*}
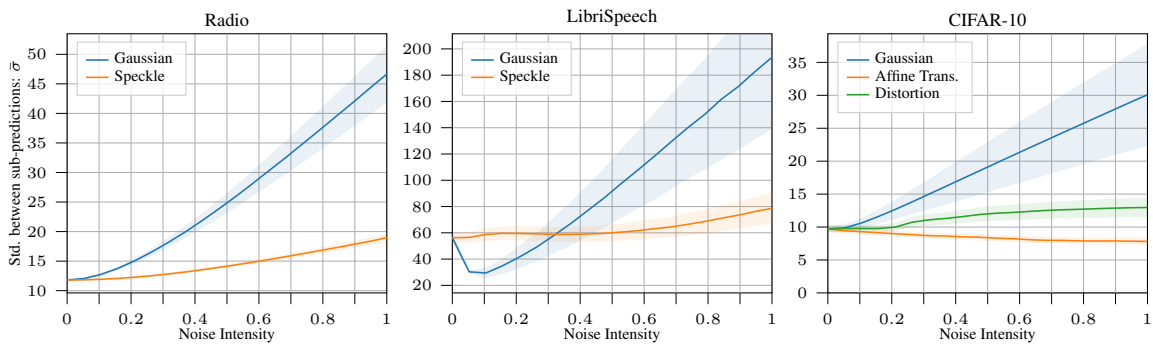

    \centering
    \input{graphs/radio/std_combined/mean_std_vs_noise_combined}
    \hspace{-1em}
    \input{graphs/libri/std_combined/mean_std_vs_noise_combined}
    \hspace{-1em}
    \input{graphs/cifar/std_combined/mean_std_vs_noise_combined}
    \caption{Radio, LibriSpeech and, CIFAR-10: mean standard deviation ($\bar{\sigma}$) across increasingly higher noise intensities.}\label{fig:std_vs_noise_radio_libri_cifar}
\end{figure*}

\begin{figure*}[ht]
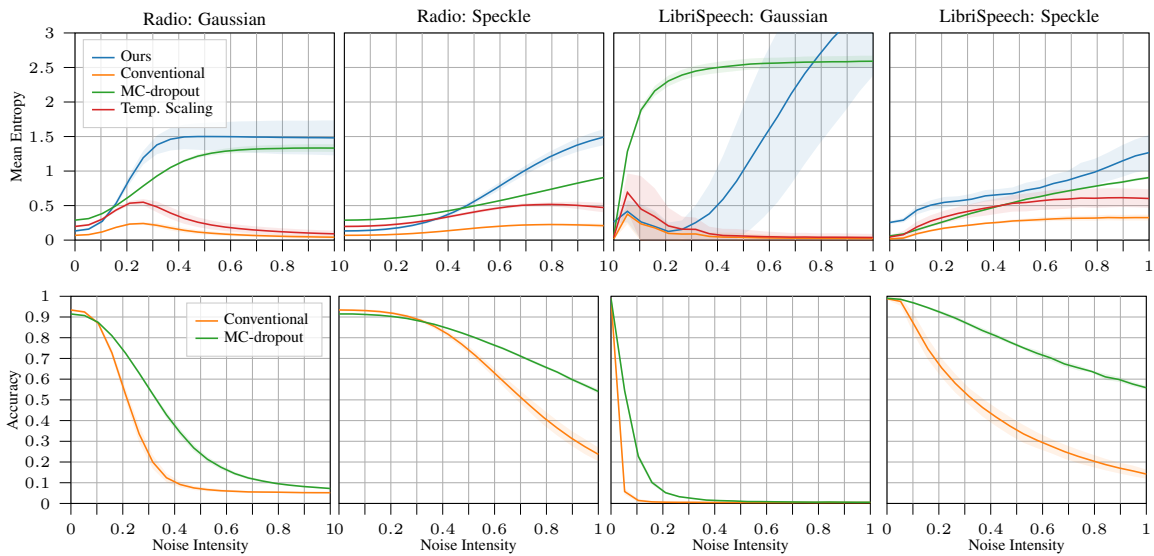

	\centering
	\hspace{0.025em}
	\input{graphs/radio/gauss/mean_entrop_vs_noise_combined}%
	\hspace{-1.5em}%
	\input{graphs/radio/speckle/mean_entrop_vs_noise_combined}%
	\hspace{-1.5em}%
	\input{graphs/libri/gauss/mean_entrop_vs_noise_combined}
	\hspace{-1.5em}%
	\input{graphs/libri/speckle/mean_entrop_vs_noise_combined}\\
	\input{graphs/radio/gauss/acc_vs_noise_combined}%
	\hspace{-1.8em}
	\input{graphs/radio/speckle/acc_vs_noise_combined}
	\hspace{-1.9em}
	\input{graphs/libri/gauss/acc_vs_noise_combined}%
	\hspace{-1.5em}
	\input{graphs/libri/speckle/acc_vs_noise_combined}\\
	\caption{\textbf{Radio and LibriSpeech}: Mean entropy and their respective accuracies across different noise intensities. Our approach and Temperature Scaling have the same accuracy as conventional.}
	\label{fig:entrop_datasetshift_radio_libri}
\end{figure*}

\begin{figure*}[ht]
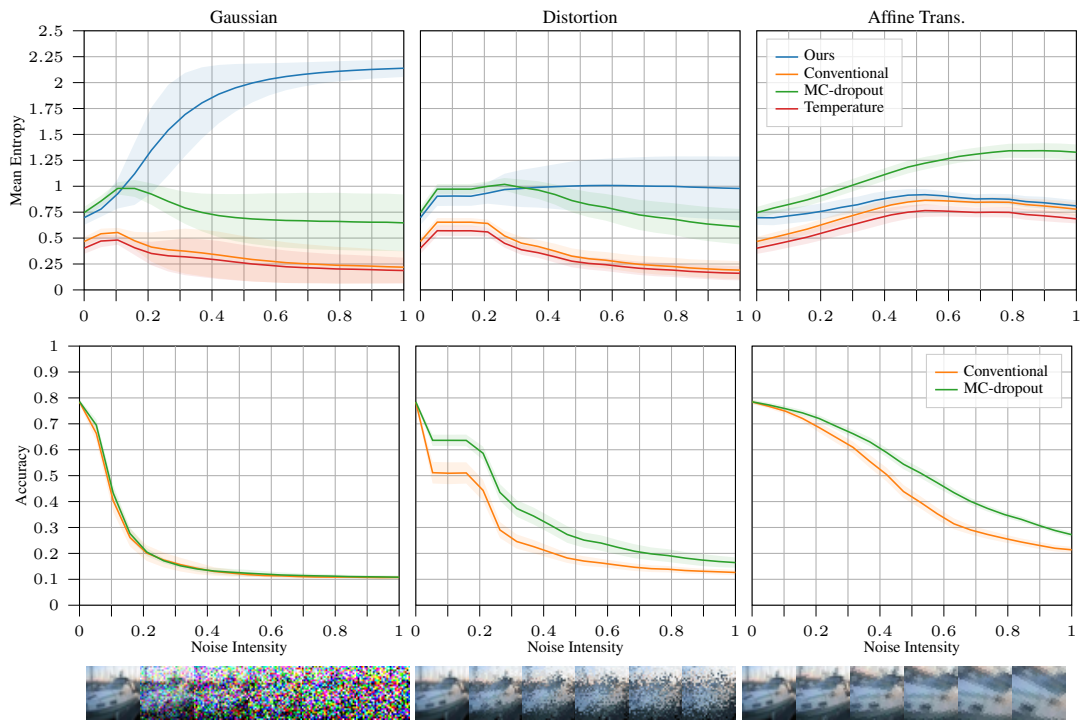

	\centering
	\input{graphs/cifar/gauss/mean_entrop_vs_noise_combined}%
	\hspace{-1.5em}
	\input{graphs/cifar/distortion/mean_entrop_vs_noise_combined}%
	\hspace{-1.5em}
	\input{graphs/cifar/affine_trans_zoom/mean_entrop_vs_noise_combined} \\
	%
	%
	\input{graphs/cifar/gauss/acc_vs_noise_combined}%
	\hspace{-1.5em}
	\input{graphs/cifar/distortion/acc_vs_noise_combined}%
	\hspace{-1.5em}
	\input{graphs/cifar/affine_trans_zoom/acc_vs_noise_combined} \\
	%
	%
	\hspace{2em}
	\input{graphs/cifar/examples/noisy_images_2_gaussian/img} \hspace{-0.5em}
	\input{graphs/cifar/examples/noisy_images_2_elastic_distortion/img} \hspace{-0.5em}
	\input{graphs/cifar/examples/noisy_images_2_affine_transform/img} \\

	\caption{\textbf{CIFAR-10}: Mean entropy and their respective accuracies across different noise intensities. Our approach and Temperature Scaling have the same accuracy as conventional.}
	\label{fig:entrop_datasetshift_cifar}
\end{figure*}

The results in Fig.~\ref{fig:std_vs_noise_radio_libri_cifar} show that on the Radio dataset, as noise increases (and accuracy decreases), $\bar{\sigma}$ increases monotonically at two different rates, in line with the accuracy drop for each noise type (Fig.~\ref{fig:entrop_datasetshift_radio_libri}~and~\ref{fig:entrop_datasetshift_cifar}). This suggests that variance between sub-patch predictions is a reliable uncertainty proxy in this setting.

For LibriSpeech, while speckle noise produces $\bar{\sigma}$ values that increase proportional to the model's accuracy drop (Fig.~\ref{fig:entrop_datasetshift_radio_libri}), Gaussian noise shows an initial dip before recovering, which is suboptimal. The cause of this dip remains unclear. We initially hypothesized that it was due to the number of classes (251) and the nonlinear punishment of softmax in cross-entropy loss on logits, which can cause large variance values for certain logits. For instance, consider two sets of predicted logits: $\left(\begin{smallmatrix}
                                                                                                                                                                                                                                                                                                                                                                                                                                                                                                                                                                             17 & 0.05 & 0.01& -0.05
\end{smallmatrix} \right)$ and $\left(\begin{smallmatrix}
                                          17 & -20 & -20& -20
\end{smallmatrix} \right)$, where the first index corresponds to the correct label. Despite their large numerical differences, these logits produce similar cross-entropy losses because the softmax function pushes the negative and near 0 values toward zero. As a result, the variance values $\sigma_k$ for $2 \le k \le 4$ become high. However, training on a 10-class subset resulted in similar behavior, suggesting that class count is not the root cause. We also tested whether sub-patch length played a role, but experiments with varying context windows and filtering for long audio samples all preserved this initial dip.

For CIFAR-10, the method generalizes surprisingly well despite the weak informative sub-patches assumption (evidenced by an average sub-patch accuracy of $0.27 \pm 0.013$). The variances growths observed for Gaussian and distortion noise are proportional to their respective accuracy drops (Fig.~\ref{fig:entrop_datasetshift_cifar}), indicating potential applications beyond the assumed dataset assumptions.
However, the affine transformation noise results in a negative correlation, which shows an important limitation: sub-patches must be sufficiently different from each other for variance to serve as a reliable proxy. Otherwise, variance remains artificially low despite high uncertainty.

\textbf{Predictive Entropy}:
We now analyze confidence across methods using predictive entropy under dataset shift (Radio and LibriSpeech: Fig.~\ref{fig:entrop_datasetshift_radio_libri}, and CIFAR-10: Fig.~\ref{fig:entrop_datasetshift_cifar}).

Across all datasets, both the conventional NN and Temperature-Scaling show some uncertainty at low noise levels, as seen in their increasing entropy. However, at higher noise levels, both methods becomes overly confident despite being incorrect, showing the need for better post-hoc recalibration.

On the Radio dataset, Variance-based Smoothing produces entropy trends that align well with accuracy drops, performing on par with MC-dropout across both noise types.

For LibriSpeech, despite the initial drop in standard deviation (Fig.~\ref{fig:std_vs_noise_radio_libri_cifar}), Variance-based Smoothing still improves uncertainty estimates over the conventional NN, particularly at higher Gaussian noise levels. Importantly, due to the max operation in Eq.~\ref{eq:sigma_bar_and_sigma_tilde}, our approach suffers less in regions where variance is less reliable. For Speckle noise, Variance-based Smoothing provides consistently increasing entropy levels, whereas the conventional model flattens around a noise level of 0.6. Nonetheless, MC-dropout remains the better choice for LibriSpeech.

On CIFAR-10, entropy results are similar to those in the Radio dataset. Variance-based Smoothing significantly improves uncertainty estimates for Gaussian and distortion noise, even outperforming MC-dropout. Interestingly, MC-dropout shows an entropy drop at higher noise levels, resulting in varying entropy scores. This suggests that some runs are more confident under noise than on clean data, an undesirable outcome.
Continuing with Variance-based Smoothing, the earlier observed negative correlation under affine transformation noise (Fig.~\ref{fig:std_vs_noise_radio_libri_cifar}) is tempered by the max and $\beta$-shift operations in Eq.~\ref{eq:sigma_bar_and_sigma_tilde}, resulting in entropy behavior similar to that of a conventional neural network.

\subsection{Computational Cost}
Table~\ref{tbl:computational_cost} compares inference costs on CIFAR-10 in terms of peak memory usage and approximate FLOPs. We use the same parameters as in our previous CIFAR-10 experiments. For MC-dropout, $n=10$ samples are processed in parallel in a single forward pass with batch size $\text{128} \cdot \text{n}$. We include an ensemble as an additional reference with  $M=10$ models loaded simultaneously.
In the table, $C_\text{fwd}$ represents the total FLOPs for one forward pass, and $C_\text{std}$ is the small overhead of computing the standard deviations across the logits.

MC-dropout and ensembles require multiple forward passes, resulting in higher computational costs. MC-dropout requires $n$ times more FLOPs and 11.5 times more memory than the conventional NN, while the ensemble requires $M$ times more FLOPs and 9.5 times more memory. In contrast, Variance-based Smoothing adds only a small overhead for variance calculations, keeping both FLOPs and memory requirements close to the conventional NN. This makes it especially suitable for real-time applications.

\begin{table}
    \centering
    \caption{Maximum memory usage (GB) and FLOPs during a single forward call on a 128-sized batch of $32\times32$ RGB images with the architecture used for the CIFAR-10 experiments.} \label{tbl:computational_cost}
    \begin{tabular}{lll}
        \toprule 
        \bfseries Method             & \bfseries Memory (GB) & \bfseries FLOPs               \\
        \midrule 
        \textbf{Conventional}        & 0.35 $\pm$ 0.00       & $C_\text{fwd}$                \\
        \textbf{Temp. Scaling}       & 0.35 $\pm$ 0.00       & $C_\text{fwd}$                \\
        \textbf{Ensemble} ($M=10$)   & 3.34 $\pm$ 0.14       & $M \cdot C_\text{fwd}$        \\
        \textbf{MC-dropout} ($n=10$) & 4.04 $\pm$ 0.00       & $n \cdot C_\text{fwd}$        \\
        \midrule
        \textbf{Ours}                & 0.35 $\pm$ 0.06       & $C_\text{fwd} + C_\text{std}$ \\
        \bottomrule 
    \end{tabular}
\end{table}

\subsection{Variance-based Smoothing as an Extension of Ensembles}\label{sec:experim_var_based_smooth_ensemb}
Continuing the discussion on the flexibility of ensemble distributions in large-class-count datasets (Sec.~\ref{sec:meth_ensemb_extension}), we empirically study the behavior of these ensembles on LibriSpeech in comparison to Variance-based Smoothing as an ensemble extension. In our approach, the variance is thus computed across logits from different models rather than sub-patches. We repurpose the conventional NNs' weights from the LibriSpeech experiments (\ref{sec:experim_setup}) as an ensemble with $M=10$ models. For the Variance-based Smoothing ensemble, we set $\alpha=1$ and $\beta$ to the negative 75th percentile of all $\bar{\sigma}$ values computed across the validation set.
\begin{figure}
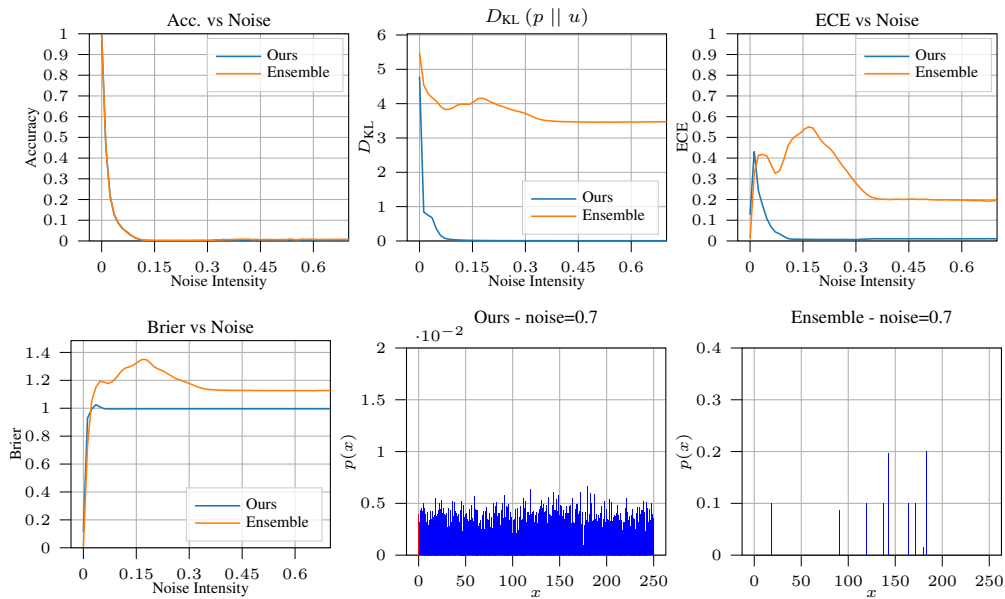

    \centering
    \input{graphs/ensemb_flexibil_libri/accuracy_vs_noise_combined}
    \hspace{-0.25cm}
    \input{graphs/ensemb_flexibil_libri/kl_vs_noise_combined}
    \hspace{-0.25cm}
    \input{graphs/ensemb_flexibil_libri/ece_vs_noise_combined} \\
    \input{graphs/ensemb_flexibil_libri/brier_vs_noise_combined}
    \hspace{-0.25cm}
    \input{graphs/ensemb_flexibil_libri/distrs/ensemb_softscale_amp=1_models=10/distribs_under_noise_sample=4/distr_noise_59}
    \hspace{-0.39cm}
    \input{graphs/ensemb_flexibil_libri/distrs/ensemb_avg_models=10/distribs_under_noise_sample=4/distr_noise_59} \\
    \caption{Comparison of ensemble flexibility on LibriSpeech under increasing Gaussian noise. The two figures on the bottom right depict a single predicted softmax vector on an input with noisy intensity of 0.7.}
    \label{fig:ensemb_flexibil_libri_combined_graphs}
\end{figure}
The results in Fig.~\ref{fig:ensemb_flexibil_libri_combined_graphs} show that accuracy deteriorates at a similar rate for both methods as Gaussian noise increases. However, key differences appear in uncertainty estimation and calibration:

The KL-divergence with the uniform distribution starts high for both methods, indicating high confidence on the clean set. As noise increases and accuracy drops, KL-divergence decreases, as expected. However, the ensemble stagnates at KL-divergence around 3.5, failing to approach the optimal 0. In contrast, Variance-based Smoothing continues to decrease smoothly to 0, without the undesired fluctuations seen in the ensemble.

The expected calibration error (ECE), which measures the weighted average difference between predicted confidence and actual accuracy across multiple confidence bins (0 being optimal), is initially lower for ensembles due to their high-confidence predictions frequently being correct. However, as noise increases beyond a certain threshold (approximately 0.05), Variance-based Smoothing significantly outperforms ensembles, achieving near-zero calibration errors at higher noise levels.

This discrepancy is further illustrated by the two examples showing an ensemble's predicted distribution compared to ours on a distorted input with Gaussian noise = 0.7. The ensemble produces a few sharp peaks but fails to approximate the required uniform distribution over 251 classes and assigns 0 probability to the true class. Variance-based Smoothing, however, generates a near-uniform distribution while still assigning some probability to the true class label (marked in red, first column of the distribution's figure).

Additionally, the Brier score, which captures both calibration and sharpness of probabilistic predictions~\citep{degroot1983comparison}, shows that Variance-based Smoothing generally achieves lower values under dataset shift, indicating better overall uncertainty estimation.

These results suggest that small ensembles ($M=10$) lack the expressiveness needed to represent a uniform distribution in high-class-count settings. Variance-based Smoothing provides a viable alternative, achieving more flexible uncertainty estimates with minimal computational overhead.

    \section{Related Work}

\textbf{Ensemble-based and Efficient Single-Model UQ Approaches}:
Deep Ensembles~\citep{DBLP:conf/nips/Lakshminarayanan17} remain a gold standard for UQ due to their robust performance under dataset shift~\citep{DBLP:conf/nips/SnoekOFLNSDRN19}. However, they require multiple forward passes and substantial memory, limiting their practicality, especially for edge applications. Extensions like Deep Sub-Ensembles~\citep{valdenegro_deep_sub_ensembles} reduce computational costs by training subsets of models but still require significant memory overhead.

Several single-model approaches have been developed. MC Dropout~\citep{DBLP:conf/icml/GalG16} approximates Bayesian inference via stochastic forward passes. SWA-Gaussian~\citep{DBLP:conf/nips/MaddoxIGVW19} produce uncertainty estimates from a single model by weight averaging over multiple training checkpoints, while Spectral-normalized Neural Gaussian Processes (SNGP)~\citep{liu_uncertainty_via_distance_awareness} achieves this through distance-awareness. Other approaches, such as Simultaneous Quantile Regression (SQR) and Orthonormal Certificates (OCs)~\citep{tagasovska_single_model_uncertainties} and Evidential Deep Learning (EDL)~\citep{sensoy_evidential_deep_learning}, modify the training objective to directly predict uncertainty. Although these methods provide viable alternatives, they typically require either multiple forward passes over the full input or significant modifications during training.

\textbf{Post-hoc Calibration Methods}:
Post-hoc methods aim to improve calibration without changing the model architecture or training process. Temperature scaling~\citep{pmlr-v70-guo17a} learns a single temperature parameter on a validation set to rescale softmax outputs, thereby improving calibration. Recent extensions, such as Dirichlet calibration~\citep{kull_beyond_temperature_scaling} and Bayesian Binning into Quantiles (BBQ)~\citep{mahdi_calibrated_probabilities_using_bayesian_binning}, allow for flexible adjustment of predictive distributions. In contrast, our method of softmax scaling is based on the variance between sub-predictions, capturing uncertainty dynamically rather than relying on a fixed set of parameters such as these approaches.

Test-time augmentation methods for UQ~\citep{ayhan2018test, DBLP:conf/sdm/JiangZWZ22, DBLP:conf/aistats/WuW24} improve uncertainty estimates by generating multiple predictions using data augmentations such as cropping and rotation. While these approaches also rely on data redundancy, they use artificial transformations rather than naturally occurring redundancy in the data and still require multiple forward passes. Among existing methods, this is the most closely related to exploiting the structure of the input data for UQ.

    \section{Discussion}
Our main finding is that variance between sub-predictions can serve as a fast and low-cost proxy for uncertainty across different tasks and noise settings. Interestingly, our results indicate that this metric may be viable beyond tasks that strictly satisfy the informative sub-patches assumption, suggesting potential for broader applicability. When used as a post-hoc recalibration method through Variance-based Smoothing, uncertainty estimates improve significantly on both clean and noisy data. In many cases, it performs on par with, and occasionally surpasses, MC-dropout, despite being considerably less computationally expensive. Additionally, applying Variance-based Smoothing as an extension to ensembles by using variance between model predictions improves their flexibility and leads to lower expected calibration errors in high-class-count settings.

\textbf{Limitations and Future Work}: While Variance-based Smoothing using variance from sub-predictions shows strong results, its effectiveness appears somewhat dataset- and noise-dependent. Future work should further investigate its performance across a wider range of datasets and noise types to better understand its limitations and generalization capabilities. Since our experiments focus solely on classification in ConvNet-based architectures, generalization to other architectures and machine learning tasks should be explored.

Our approach relies on standard deviation across logits as a measure of disagreement between sub-predictions. However, this is a relatively simple metric, and the softmax function's exponential nature causes logits to be transformed nonlinearly, which can impact variance estimation. Future research could explore alternatives that better account for this non-linearity and potentially improve uncertainty estimates.

Lastly, Variance-based Smoothing computes variance on the model’s final-layer outputs. Future work could explore whether intermediate feature activations provide additional uncertainty information. Easier-interpretable latent space regularizers, such as~\citet{DBLP:journals/corr/abs-2408-12936, burgess_beta_vae}, may offer a promising starting point for studying this behavior.

   	\section*{Acknowledgments}
   	I thank Thayheng Nhem for the discussions that led to the main idea of this work. His insights, especially in the early stages, were very helpful. This research is funded by the University of Antwerp's Department of Computer Science.

    \bibliography{biblio}
    \newpage

\onecolumn

\title{Efficient Post-Hoc Uncertainty Calibration via Variance-Based Smoothing\\(Supplementary Material)}
\maketitle
\appendix

\section{Experimental Setup} \label{apdx:experimental_setup}
All single-model experiments are repeated 10 times with unique random seeds.

\subsection{Radio}
We follow \citet{NhemDWPOP25} unless stated otherwise. The model is trained in a fully supervised manner using the Adam optimizer for 30 epochs with a learning rate of $1 \times 10^{-3}$ and a batch size of 32. Training is performed using cross-entropy loss.

The model architecture, shown in Table~\ref{tab:architecture_radio}, is identical to \citet{NhemDWPOP25}, except that the GRU layer is omitted for simplicity. Batch normalization is applied after each convolution layer, followed by a ReLU activation, except for the final layer, where neither batch normalization nor ReLU is applied.

The dataset follows the train-validation-test split used by \citet{NhemDWPOP25}, with 70,000 training examples (70\%), 9,900 validation examples (10\%), and 20,100 test examples (20\%). To improve consistency between sub-predictions, the first 500 samples of each example are removed. As a result, predictions are made on $1580\times 2$ inputs, resulting in a $79\times20$ tensor before the final average pooling layer. The model has a total downscaling factor of 20.

\begin{table}[h]
    \caption{Architecture for the Radio experiments. ${}^a$Variance is computed across logits from this layer.}
    \centering
    \begin{tabular}{lcccc}
        \toprule
        \textbf{Layer} & \textbf{Output}  & \textbf{Kernel} & \textbf{Stride} & \textbf{Padding} \\
        \midrule
        Input          & $1580 \times 2$  &                 &                 &                  \\
        Conv           & $315 \times 512$ & 10              & 5               & 2                \\
        Conv           & $78 \times 512$  & 8               & 4               & 2                \\
        Conv           & $79 \times 512$  & 4               & 1               & 2                \\
        Conv           & $80 \times 512$  & 4               & 1               & 2                \\
        Conv           & $79 \times 512$  & 4               & 1               & 1                \\
        \midrule
        Conv           & $79 \times 20^a$ & 1               & 1               & -                \\
        Avg. pool      & $1 \times 20$    & -               & -               & -                \\
        \bottomrule
    \end{tabular}
    \label{tab:architecture_radio}
\end{table}

\subsection{LibriSpeech}
The protocol follows that of the fully supervised LibriSpeech models described in \citet{DBLP:conf/nips/LoweOV19, DBLP:journals/corr/abs-2408-12936}. The architecture, shown in Table~\ref{tab:architecture_libri}, is the same, except that the GRU layer is removed, and batch normalization is added after each convolution layer, except for the final one.

The model is trained using the Adam optimizer with a learning rate of $2 \times 10^{-4}$ for 1,000 epochs with a batch size of 8.

The dataset consists of spoken audio sampled at 16 kHz. Since \citet{DBLP:conf/nips/LoweOV19} does not explicitly provide a validation set, we split the test in half with 50\% for validation and 50\% for testing, resulting in 22,830 training examples (80\%), 2,854 validation examples (10\%), and 2,854 test examples (10\%). As examples contain variable-length audio, random crops of length 20,480 (1.28s) are extracted for training and evaluation.
\begin{table}[h]
    \caption{Architecture for the LibriSpeech experiments. ${}^a$Variance is computed across logits from this layer.}
    \centering
    \begin{tabular}{lcccc}
        \toprule
        \textbf{Layer} & \textbf{Output}    & \textbf{Kernel} & \textbf{Stride} & \textbf{Padding} \\
        \midrule
        Input          & $20480 \times 1$   &                 &                 &                  \\
        Conv           & $4095 \times 512$  & 10              & 5               & 2                \\
        Conv           & $1023 \times 512$  & 8               & 4               & 2                \\
        Conv           & $512 \times 512$   & 4               & 2               & 2                \\
        Conv           & $257 \times 512$   & 4               & 2               & 2                \\
        Conv           & $128 \times 512$   & 4               & 2               & 1                \\
        \midrule
        Conv           & $128 \times 251^a$ & 1               & 1               & -                \\
        Average pool   & $1 \times 251$     & -               & -               & -                \\
        \bottomrule
    \end{tabular}
    \label{tab:architecture_libri}
\end{table}

\subsection{CIFAR-10}
We use the architecture proposed by \citet{xu2015empiricalevaluationrectifiedactivations}, which is based on the Network-In-Network (NiN) architecture~\citep{lin2013network} and provides relatively strong performance on CIFAR-10. The architecture is presented in Table~\ref{tab:architecture_cifar10}.

The model is trained using the Adam optimizer with a learning rate of $1 \times 10^{-3}$ for 250 epochs with a batch size of 128 and a weight decay of $5 \times 10^{-4}$. Note that for our MC-dropout implementations, the already present dropout layers are omitted and an additional dropout layer is added after every ReLu layer.

\begin{table}[h]
    \caption{Architecture for the CIFAR-10 experiments. ${}^a$Variance is computed across logits from this layer.}
    \centering
    \begin{tabular}{lcccc}
        \toprule
        \textbf{Layer} & \textbf{Output}           & \textbf{Kernel} & \textbf{Stride} & \textbf{Padding} \\
        \midrule
        Input          & $32 \times 32 \times 3$   &                 &                 &                  \\
        Conv           & $32 \times 32 \times 192$ & $5 \times 5$    & 1               & 2                \\
        Conv           & $32 \times 32 \times 160$ & $1 \times 1$    & 1               & 0                \\
        Conv           & $32 \times 32 \times 96$  & $1 \times 1$    & 1               & 0                \\
        Max Pool       & $16 \times 16 \times 96$  & $3 \times 3$    & 2               & 1                \\
        Dropout (0.5)  & -                         & -               & -               & -                \\
        Conv           & $16 \times 16 \times 192$ & $5 \times 5$    & 1               & 2                \\
        Conv           & $16 \times 16 \times 192$ & $1 \times 1$    & 1               & 0                \\
        Conv           & $16 \times 16 \times 192$ & $1 \times 1$    & 1               & 0                \\
        Avg. Pool      & $8 \times 8 \times 192$   & $3 \times 3$    & 2               & 1                \\
        Dropout (0.5)  & -                         & -               & -               & -                \\
        Conv           & $8 \times 8 \times 192$   & $3 \times 3$    & 1               & 1                \\
        Conv           & $8 \times 8 \times 192$   & $1 \times 1$    & 1               & 0                \\
        \midrule
        Conv           & $8 \times 8 \times 10^a$  & $1 \times 1$    & 1               & 0                \\
        Avg. Pool      & $1 \times 1 \times 10$    & $8 \times 8$    & -               & -                \\
        \bottomrule
    \end{tabular}
    \label{tab:architecture_cifar10}
\end{table}

\subsection{Noise Types} \label{apdx:different_noise_types}
Across the three datasets, inputs are incrementally distorted using different noise types with the degree of distortion controlled by $\lambda \in [0,1]$. For Radio and LibriSpeech we apply Gaussian or Speckle noise, while for CIFAR-10 we apply either Gaussian noise, affine transformations, or elastic distortions.

Gaussian noise is introduced as $\mathbf{x}_{\text{noise}} := \mathbf{x} + \lambda \mathbf{\epsilon}$, while Speckle noise is applied as $\mathbf{x}_{\text{noise}} := \mathbf{x} + \lambda (\mathbf{x} \odot \mathbf{\epsilon})$, where $\odot$ denotes element-wise multiplication. In both cases, $\mathbf{\epsilon}$ is a noise vector $\mathbf{\epsilon} \sim \mathcal{N}(\mathbf{0}, \mathbf{I})$ of the same shape as the input $\mathbf{x}$ is sampled.

In the case of an affine transformation, an image $\mathbf{X}$ of size $H\times W$ is modified through a rotation by $\theta = \lambda \cdot 30^\circ$, a shear by $s=\lambda \cdot  10^\circ$, and an isotropic scaling factor $\gamma=1+\lambda$. To prevent cropping artificats, the image is first padded by $p=0.2 \max(H, W)$ before applying the transformation, which is parameterized by the matrix
\begin{equation}
    A=\left[\begin{array}{cc}
                \gamma \cos \theta & -\gamma \sin \theta+s \\
                \gamma \sin \theta & \gamma \cos \theta+s
    \end{array}\right].
\end{equation}
Finally, the image is center-cropped back to its original dimensions to ensure that the structured perturbations induced by $\lambda$ are smoothly incorporated while maintaining spatial consistency.

When elastic distortion is applied, pixel coordinates are perturbed based on randomly generated displacement fields. Given an image $\mathbf{X}$, displacement fields $\Delta x$ and $\Delta y$ are sampled independently from a uniform distribution $\mathcal{U}(-\lambda \cdot 5, \lambda \cdot 5)$. Each pixel coordinates $(x, y)$ is displaced to
\begin{equation}
    x^{\prime}=\operatorname{clamp}(x+\Delta x, 0, W-1), \quad y^{\prime}=\operatorname{clamp}(y+\Delta y, 0, H-1),
\end{equation}
where the clamp function ensures that all coordinates remain within valid image bounds. The deformed image $\mathbf{X}'$ is then obtained by mapping each pixel to its new location using nearest-neighbor interpolation, introducing local distortions that increase in magnitude as $\lambda$ approaches 1 while preserving overall spatial coherence.

Visual examples of the noise types applied to an image are shown in Fig.~\ref{fig:entrop_datasetshift_cifar}. We also provide code implementations for all noise types in the GitHub repository.

\subsection{Baselines} \label{apdx:baselines}
We compare our method against MC-dropout and Temperature Scaling. MC-dropout is trained with the same hyperparameters as the conventional NN, except that dropout ($p=0.5$) is applied after every ReLU layer. During evaluation, dropout is maintained, and inference is performed on batches of size $(n \cdot \text{Orig. Batch Size})$, with $n = 10$.
For Temperature Scaling, we reuse the conventional NN's weights and continue training a the temperature parameter on top of these frozen weights. This scalar is initialized to 1 and optimized for 10 epochs using the Adam optimizer with a learning rate of 0.01, a batch size of 64, and Cross Entropy loss.

\section{Reliability Diagrams and Confidence Counts} \label{sec:apdx_reliabil_diagrs_and_conf_counts}

\subsection{Radio}
\begin{figure*}[h]
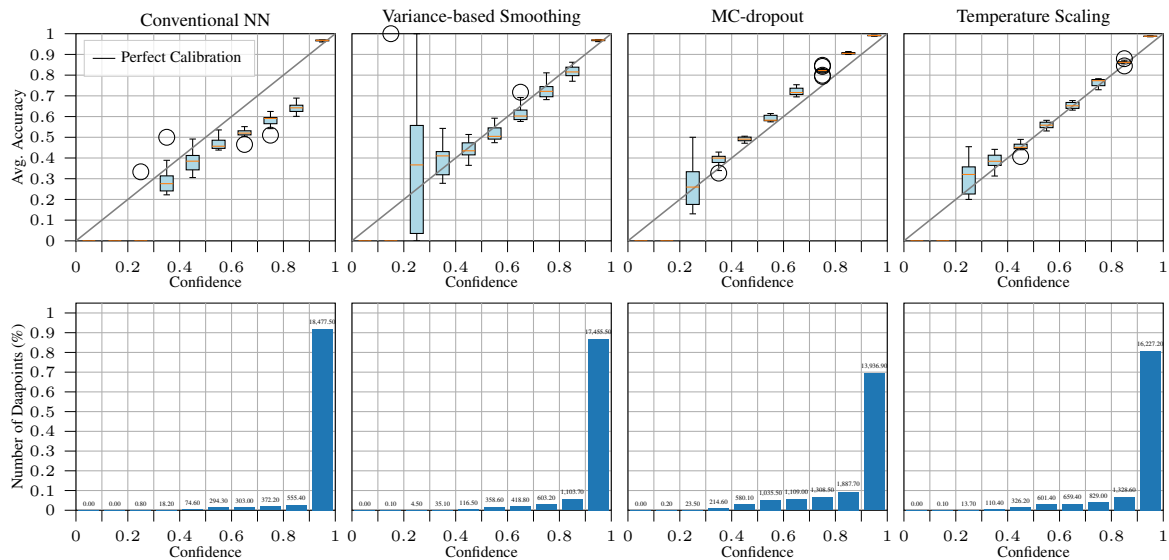

    \input{graphs/radio/reliabil_diagr/acc_vs_confidence_conventional_ep=30}\hspace{-1.5em}
    \input{graphs/radio/reliabil_diagr/acc_vs_confidence_kern=10_amp=1_auto}\hspace{-1.5em}
    \input{graphs/radio/reliabil_diagr/acc_vs_confidence_mc_samples=10_ep=30}\hspace{-1.5em}
    \input{graphs/radio/reliabil_diagr/acc_vs_confidence_temperature_post_eps=10_b=64}
    \\
    \input{graphs/radio/reliabil_diagr/datapoints_vs_confidence_conventional_ep=30}\hspace{-1.5em}
    \input{graphs/radio/reliabil_diagr/datapoints_vs_confidence_kern=10_amp=1_auto}\hspace{-1.5em}
    \input{graphs/radio/reliabil_diagr/datapoints_vs_confidence_mc_samples=10_ep=30}\hspace{-1.5em}
    \input{graphs/radio/reliabil_diagr/datapoints_vs_confidence_temperature_post_eps=10_b=64}
    \caption{\textbf{Radio}: Individual reliability diagrams and their frequency per bin (averaged over 10 runs).}
    \label{fig:apdx_radio_confidence_graphs}
\end{figure*}

\subsection{LibriSpeech}
\begin{figure*}[h]
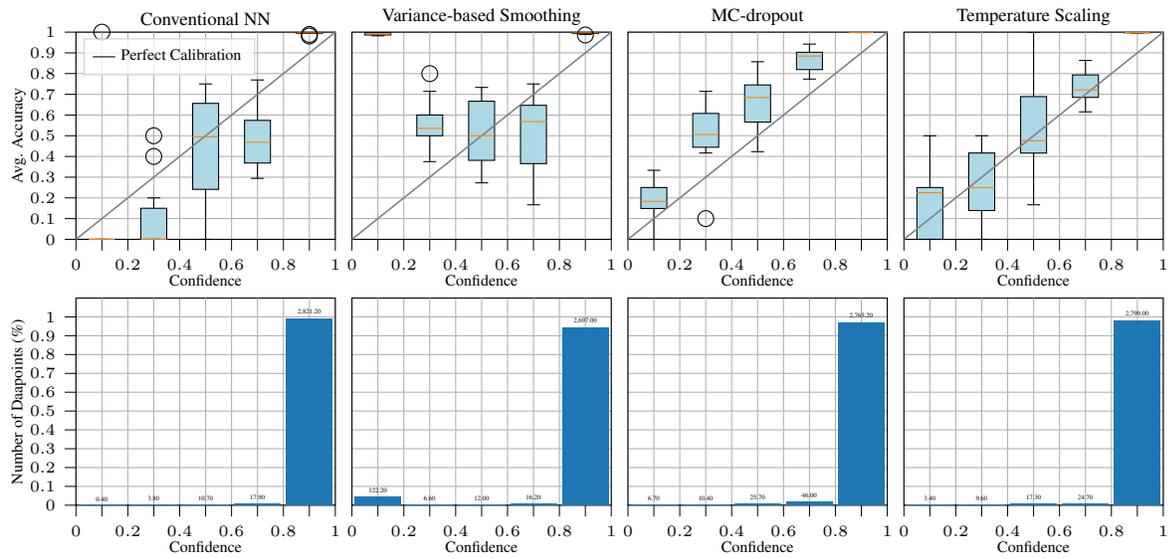

    \input{graphs/libri/reliabil_diagr/acc_vs_confidence_conventional}\hspace{-1.5em}
    \input{graphs/libri/reliabil_diagr/acc_vs_confidence_kern=40_amp=5_q95}\hspace{-1.5em}
    \input{graphs/libri/reliabil_diagr/acc_vs_confidence_mc_samples=10}\hspace{-1.5em}
    \input{graphs/libri/reliabil_diagr/acc_vs_confidence_temperature_post_eps=10_b=64}
    \\
    \input{graphs/libri/reliabil_diagr/datapoints_vs_confidence_conventional}\hspace{-1.5em}
    \input{graphs/libri/reliabil_diagr/datapoints_vs_confidence_kern=40_amp=5_q95}\hspace{-1.5em}
    \input{graphs/libri/reliabil_diagr/datapoints_vs_confidence_mc_samples=10}\hspace{-1.5em}
    \input{graphs/libri/reliabil_diagr/datapoints_vs_confidence_temperature_post_eps=10_b=64}
    \caption{\textbf{LibriSpeech}: Individual reliability diagrams and their frequency per bin (averaged over 10 runs).}
    \label{fig:apdx_libri_confidence_graphs}
\end{figure*}

\subsection{CIFAR-10}
\begin{figure*}[h]
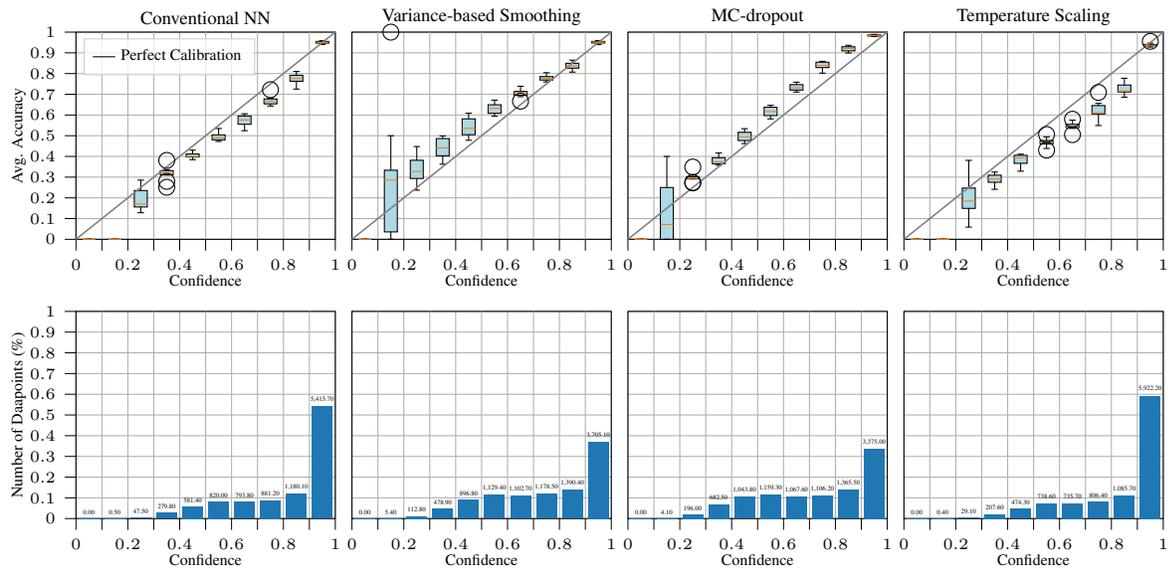

    \input{graphs/cifar/reliabil_diagr/acc_vs_confidence_conventional_ep=250}\hspace{-1.5em}
    \input{graphs/cifar/reliabil_diagr/acc_vs_confidence_kern=1_ep=250}\hspace{-1.5em}
    \input{graphs/cifar/reliabil_diagr/acc_vs_confidence_mc_samples=10_ep=250}\hspace{-1.5em}
    \input{graphs/cifar/reliabil_diagr/acc_vs_confidence_temperature_post_eps=10_b=64}
    \\
    \input{graphs/cifar/reliabil_diagr/datapoints_vs_confidence_conventional_ep=250}\hspace{-1.5em}
    \input{graphs/cifar/reliabil_diagr/datapoints_vs_confidence_kern=1_ep=250}\hspace{-1.5em}
    \input{graphs/cifar/reliabil_diagr/datapoints_vs_confidence_mc_samples=10_ep=250}\hspace{-1.5em}
    \input{graphs/cifar/reliabil_diagr/datapoints_vs_confidence_temperature_post_eps=10_b=64}
    \caption{\textbf{CIFAR-10}: Individual reliability diagrams and their frequency per bin (averaged over 10 runs).}
    \label{fig:apdx_cifar_confidence_graphs}
\end{figure*}

\end{document}